\documentclass[review]{elsarticle}
\usepackage{amssymb}
\usepackage{amsmath}
\usepackage{graphicx}
\usepackage{lineno}
\usepackage{amsthm}
\usepackage{pifont}

\begin{document}
	
\ifpreprint
\setcounter{page}{1}
\else
\setcounter{page}{1}
\fi

\begin{frontmatter}
			
\title{Fourier fractal dimension to predict the generalization of deep neural networks}

\author[imecc]{Joao B. Florindo\corref{cor1}} 
\cortext[cor1]{Corresponding author}
\ead{florindo@unicamp.br}
\author[imecc]{Davi Wanderley Misturini} 
\ead{d235799@dac.unicamp.br}

\address[imecc]{Institute of Mathematics, Statistics and Scientific Computing - University of Campinas\\
Rua S\'{e}rgio Buarque de Holanda, 651, Cidade Universit\'{a}ria ``Zeferino Vaz'' - Distr. Bar\~{a}o Geraldo, CEP 13083-859, Campinas, SP, Brasil}

\begin{abstract}
Predicting the generalization performance of deep neural networks without relying on hold-out validation data is a fundamental challenge in machine learning. While Stochastic Gradient Descent (SGD) drives the optimization of these highly parameterized models, its heavy-tailed, non-Gaussian dynamics induce complex, scale-invariant trajectories in the parameter space. In this paper, we propose a novel generalization measure based on the Fourier fractal dimension of the network's weight variations. By analyzing the characteristic function of the Lévy-driven stochastic differential equations in the frequency domain, we extract a metric that robustly captures the geometric complexity of the learning process. Furthermore, we introduce a customized Fourier-based optimizer designed to actively regularize this fractal dimension during training. Extensive empirical evaluations on the CIFAR-10, SVHN, and MNIST datasets demonstrate that our proposed Fourier generalization measure exhibits a strong correlation with the actual generalization gap. Our method achieves state-of-the-art Kendall rank correlation coefficients, outperforming a wide array of existing norm-based, margin-based, and PAC-Bayesian measures. Ultimately, this work highlights the potential of frequency-domain fractal analysis as both a powerful predictor for model generalizability and a principled foundation for developing more stable optimization algorithms.
\end{abstract}

\end{frontmatter}

\section{Introduction}

Predicting the generalization performance of a deep neural network without using any external data is a fundamental task in machine learning. A well-succeeded approach to this problem brings several benefits. For example, it eliminates the need to remove part of the annotated data from the training set to be used in the validation/test set. Besides the possibility of that removed data being biased in some sense, that extra data might be useful for better training performance and even to improve generalization, as it is well known that using larger training datasets prevents overfitting. Predicting generalization with robustness is also useful in autoML, in the automatic search for the best architecture and hyperparameters for a specific task. Techniques to predict the generalization also provide a valuable direction to better understanding of the learning process as, by investigating how the generalization can be predicted, we are also understanding which aspects play more important role in the learning algorithm. And an immediate and quite practical consequence of better understanding the learning process is the developed of better learning algorithms, more focused on the generalization capability, which is at the end the most important points in real-world applications.

\section{Related Works}

Predicting the generalization gap of deep neural networks has spurred a diverse range of theoretical and empirical investigations. Traditional statistical learning theory often yields vacuous bounds for heavily overparameterized networks \cite{zhang2017understanding}. Consequently, recent literature has shifted towards complexity measures based on the properties of the learned weights and the geometry of the loss landscape. Prominent approaches include norm-based bounds \cite{neyshabur2015norm}, margin-based analysis \cite{bartlett2017spectrally}, and PAC-Bayesian frameworks \cite{dziugaite2017computing}. In a comprehensive empirical study, \cite{jiang2019fantastic} evaluated dozens of these generalization measures, revealing that while some PAC-Bayes and margin-based metrics show promise, there remains a significant need for measures that consistently correlate with the generalization gap across varying architectures and datasets without relying on validation data.

Recently, the literature has explored the use of fractal geometry and related concepts to analyze and measure the generalization capacity of deep neural networks. This line of research is largely motivated by the empirical observation that the trajectories of iterative optimization algorithms, such as stochastic gradient descent (SGD), frequently exhibit complex fractal structures \cite{simsekli2020hausdorff}. The authors established a formal connection between the generalization error and the Hausdorff dimension of the heavy-tail driven weight distribution.

Expanding this perspective to bypass the curse of dimensionality, \cite{birdal2021intrinsic} proposed the use of tools from Topological Data Analysis (TDA) to compute the intrinsic dimension via persistent homology. They demonstrated that this topological measure acts as a strong practical predictor of generalization, without requiring strict statistical or geometric assumptions about the training dynamics.

To address the strong dependence on Lipschitz continuity assumptions required in previous analyses, which often do not hold in modern neural networks, \cite{dupuis2023generalization} introduced generalization bounds grounded in data-dependent fractal dimensions. By establishing the notion of geometric stability, they managed to control the generalization error using mutual information terms in conjunction with fractal geometry. More recently, direct empirical measures that estimate the fractal dimension of the parameter space through local perturbations have also been proposed \cite{florindo2024empirical}, achieving competitive results in predicting the generalization gap with significantly lower computational cost, as they do not require the entire optimization trajectory.

\section{Background}

\subsection{Fractal Dimension}

The concept of fractal dimension \cite{falconer2013fractal} extends the traditional notion of topological dimension to quantify the complexity and space-filling capacity of a geometric object, signal, or data manifold. While several mathematical definitions exist, the box-counting dimension (also known as Minkowski-Bouligand dimension) is particularly prominent due to its empirical tractability. For a given set $S$ in a metric space, let $N(\epsilon)$ be the minimum number of boxes of side length $\epsilon$ required to completely cover $S$. The box-counting dimension $D_{box}$ is defined as
\begin{equation}
D_{box} = \lim_{\epsilon \to 0} \frac{\log N(\epsilon)}{\log(1/\epsilon)}.
\end{equation}
In the context of machine learning, fractal dimension has been utilized to characterize the intrinsic dimensionality of data representations and the complex trajectories traversed by optimization algorithms during training.

\subsection{Fourier Fractal Dimension}

The Fourier fractal dimension provides a frequency-domain perspective on the complexity of a signal or a high-dimensional parameter landscape \cite{falconer2013fractal}. For a fractal profile, the power spectral density $P(\omega)$ typically follows a power-law relationship with respect to the frequency $\omega$, expressed as
\begin{equation}
P(\omega) \propto |\omega|^{-\beta},
\end{equation}
where $\beta$ is the spectral exponent. By extracting the slope from the log-log plot of the power spectrum against frequency, one can directly relate $\beta$ to the fractal dimension of the profile. In our framework, we leverage this mathematical relationship by analyzing the Fourier transform of the network's weight variations, effectively capturing the multi-scale structural complexity of the parameters induced by the stochastic nature of the training dynamics.

\subsection{Mathematical Formulation of the Generalization Problem}

The training algorithm of supervised tasks in machine learning can be generally formulated as an unconstrained and non-convex optimization problem:
\begin{equation}
\label{eq:optimization}
\min_{w\in\mathbf{R}^d}\left( f(w) := \frac{1}{n}\sum_{i=1}^n f^{(i)}(w) \right), 
\end{equation}
where $f$ is the objective cost function, $f^{(i)}$ is the value of $f$ for a single data point indexed by $i$, $n$ is the number of data points, and $w$ is the $d$-length vector of parameters to be adjusted (learned).

Stochastic Gradient Descent (SGD) is probably the most popular approach to solve this problem in deep learning. This is an iterative numerical method whose iteration is formulated as
\begin{align}
    \begin{split}
    \nabla\tilde{f}_k(w):=\frac{1}{B}\sum_{i\in\tilde{B}_k}\nabla f^{(i)}(w),\\
    w_{k+1} = w_k - \eta\nabla \tilde{f}_k(w_k),
    \end{split}
\end{align}
where $\tilde{B}_k$ is the \textit{batch}, a random subset of $\{1,2,\cdots,n\}$ with cardinality $B$.

Here, we focus on the generalization problem from the perspective of a classification task, even though the extension to other domains is relatively straightforward. Let $\mathcal{D}$ represent the distribution from which we draw $m$ i.i.d. tuples to compose a dataset $\mathcal{S} = \{(X_1,y_1),\cdots,(X_m,y_m)\}$, where $X_i$ denotes the input data and $y\in\{1,\cdots,\kappa\}$ is the corresponding label. Our deep neural network can be represented by $f(w)$ satisfying (\ref{eq:optimization}), but now we reformulate our notation to make its dependence on $X$ explicit. More exactly, we denote the $k^{th}$ output (class) of the network parameterized by $w$ for the input $X$ by $f_w(X)[k]$. Our loss function is therefore the `0-1' loss. Over the data distribution, we define the expected loss $L(f_w) = \mathbb{E}_{(X,y)\sim\mathcal{D}}\left[ I(f_w(X)[y]\leq \max_{j\neq y}f_w(X)[j]) \right]$, where $I$ is the indicator function. Similarly, we define the empirical loss over $\mathcal{S}$ by $\hat{L}(f_w) = \frac{1}{m}\sum_{i=1}^{m}\left[ I(f_w(X)[y_i]\leq \max_{j\neq y_i}f_w(X)[j]) \right]$. The \textit{generalization gap} is defined by the difference $L(f_w)-\hat{L}(f_w)$.

\section{Proposed Method}

\subsection{Theoretical Motivation}

A well established research line on the comprehension of SGD dynamics consists of analyzing statistical properties of the parameters $w$ \cite{zhu2019anisotropic,simsekli2019tail}. Such approach mostly relies on viewing SGD as a discretization of a stochastic differential equation (SDE). While a typical assumption in this framework is that the gradient noise $\nabla\tilde{f}_k(w)-\nabla f(w)$ obeys a Gaussian distribution, more recent studies have questioned this point \cite{gurbuzbalaban2021heavy} and verified that, in practice, this distribution is characterized by heavy tail. In this scenario, SGD is viewed as the Euler-Maruyama discretization of the following SDE:
\[ dW_t = -\nabla f(W_t)dt + \Sigma_1(W_t)dB_t + \Sigma_2(W_t)dL_t^{\alpha(W_t)}, \]
where $W_t$ is the stochastic process modeling the evolution of $w$ over time $t$ (which is an abstraction of the training epoch), $\Sigma_1$, $\Sigma_2$ are $d \times d$ matrix-valued functions, and $L_t^{\alpha}$ is the state-dependent $\alpha$-stable Lévy motion. This is a quite general representation that encompasses many complex optimization algorithms. This includes, for example, the modeling of SGD as an Ornstein-Uhlenbeck process when $\Sigma_2 = 0$ \cite{zhu2019anisotropic} and the isotropic configuration in \cite{simsekli2019tail} when $\Sigma_2$ is diagonal and $\alpha_i(w) = \alpha \in (0, 2]$ for all $i, w$.

\begin{figure}
    \centering
    \begin{tabular}{cc}         
    Layer 1 & Layer 6\\
    \includegraphics[width=0.45\textwidth]{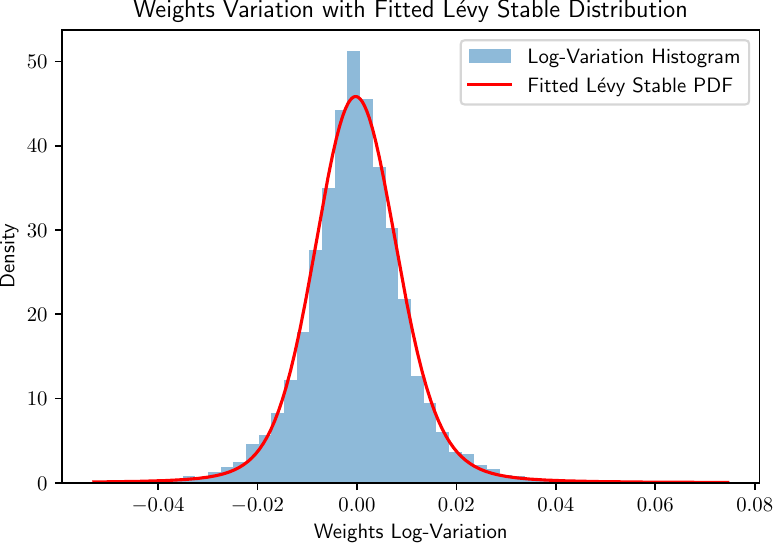} &
    \includegraphics[width=0.45\textwidth]{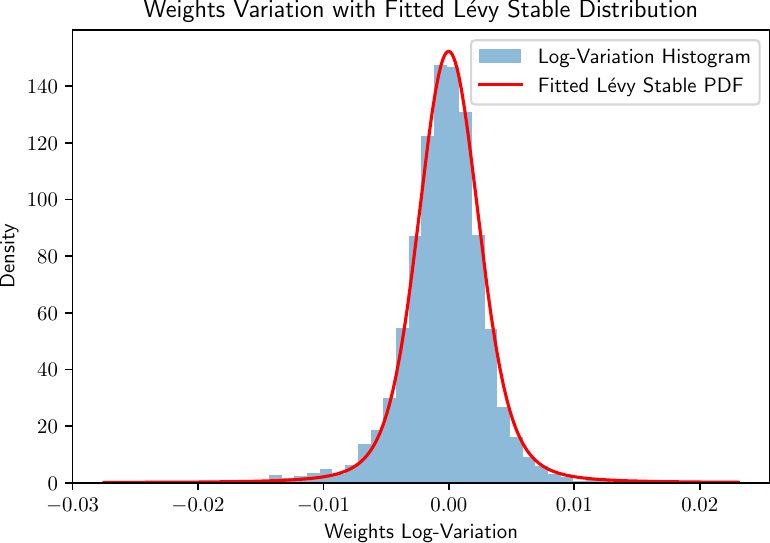}\\
    Layer 11 & Layer 16\\
    \includegraphics[width=0.45\textwidth]{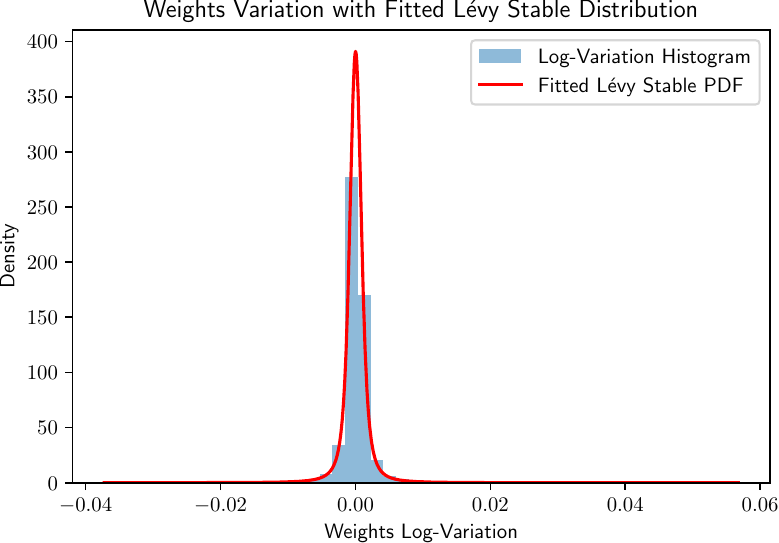} &
    \includegraphics[width=0.45\textwidth]{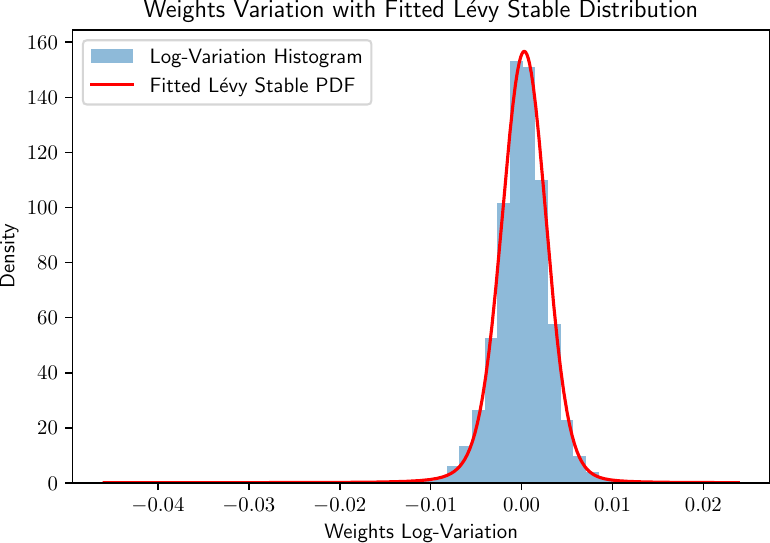}
    \end{tabular}
    \caption{Lévy stable distribution probability density function (PDF) fitted to the histogram of weights log-variation.}
    \label{fig:levypdf}
\end{figure}

An important characterization of Levy-driven processes is provided by its charactetistic funcion, whose analytical expression by given by the Levy–Khintchine formula:
\begin{equation}
\label{eq:chf}
\varphi_{W_t}(u) = \mathbb{E}\left[ \mathbf{e}^{iuW_t} \right] = \exp\left( t \left( aiu-\frac{1}{2}\sigma^2u^2 + \int_{\mathbb{R}\backslash\{0\}} \left( \mathbf{e}^{iuw}-1-iux\mathbf{1}_{|w|<1}\Pi(dw) \right) \right) \right), 
\end{equation}
where $a\in\mathbb{R}$ is the linear drift, $\sigma\geq 0$ relates to the Brownian motion term, and $\Pi$ is the Levy measure, which is $\sigma$-finite and satisfies
\[ \int_{\mathbb{R}\backslash\{0\}}\min(1,w^2)\Pi(dw) < \infty. \]
$\Pi(w)$ is associated with the jump component. While (\ref{eq:chf}) is hard to be interpreted and highly dependent on the training dataset, we empirically verify that considering $W_t$ as a $d$-dimensional $\alpha$-stable process still provides a reasonable approximation. Figure \ref{fig:levypdf} illustrates the Lévy stable distribution fitted to the histogram of weights log-variation, which corresponds to $\log(W_{t+1}/W_t)$ for any training epoch $t$. Here this value is computed on the training of CIFAR-10 using AlexNet for 40 epochs. We show the distribution on layers 1, 6, 11, and 16.

For such simplified version, we can take the characteristic funcion
\begin{equation}\label{eq:chf_simple} 
    \phi(u) = \exp \left( -\sigma^\alpha |u|^\alpha \left( 1-i\beta\text{sgn}(u) \tan \left( \frac{\pi\alpha}{2} \right) \right) + i\mu u \right), 
\end{equation}
where $\beta\in[-1,1]$ is the skewness parameter, which controls asymmetry, and $\mu$ is the location parameter, representing the center of the distribution.

\subsection{Proposed Generalization Measure}

The characteristic function (\ref{eq:chf_simple}) is known to be the Fourier transform of the random process $W_t$. After some straightforward algebraic manipulation, we arrive at an expression for the squared magnitude of $\phi(u)$:
\[ |\phi(u)|^2 = \exp(-2|\sigma u|^\alpha). \]
Fitting a straight line to the curve of $\log(\log(|\phi(u)|^2))$ against $\log(\sigma u)$, the slope is $\alpha$, the distribution parameter. Such relation between the logarithms of the magnitude of the Fourier transform and of the frequency is well known in fractal geometry as a mean of estimating the fractal dimension, via potential theory \cite{falconer2013fractal}.

On the other hand, previous studies have reported an interesting relation between $\alpha$ of an SGD algorithm modeled by a Levy-driven stochastic process and the ability of the objective function to jump from narrow to wider minima \cite{simsekli2019tail}. More specifically, let $f$ be $f$ be a smooth function with $r$ local minima $m_i$ and $r$ local maxima $s_i$, such that
\[ -\infty=s_0 < m_1 < s_1 < \cdots < s_{r-1} < m_r < s_r = \infty. \]
Let us further assume that $f''(m_i) > 0$ and $f''(s_i) < 0$ for all $i$, and that for large enough $|w|$ we have $f'(w) > |w|^{1+c}$ for some constant $c>0$. Under these assumptions, we define a neighborhood $B_i=\{|x-m_i|\leq\delta\}$, where $\delta>0$ should be small enough such that the neighborhood is contained in $S_i=(s_{i-1},s_i)$. For an initial point $w_0\in B_i$ we define the transition time to the neighborhood of another local minimum as $T_{w_0}^i(\epsilon) = \inf\{t\geq 0: w_t^\epsilon\notin\cup_{j\neq i}B_j\}$. A theorem demonstrated in \cite{pavlyukevich2007cooling} states that, in the limit $\epsilon\rightarrow 0$, the following holds regarding the probability of each transition time:
\begin{align}\label{eq:transition}
    \begin{split}
    &\mathbb{P}_{w_0}(T^i(\epsilon)\in B_j) \rightarrow q_{ij}q_i^{-1}, \text{ if } i\neq j,\\
    &\mathbb{P}_{w_0}(\epsilon^\alpha T^i(\epsilon)\geq u) \leq \mathrm{e}^{-q_iu}, \text{ for all } u\geq 0,
    \end{split}
\end{align}
where
\begin{align}
    \begin{split}
    &q_{ij} = \frac{1}{\alpha}\left| \frac{1}{|s_{j-1}-m_i|^\alpha}-\frac{1}{|s_j-m_i|^\alpha} \right|,\\
    &q_i = \sum_{j\neq i}q_{ij}.
    \end{split}
\end{align}
Among the interesting observations that we can obtain from (\ref{eq:transition}), we have that, in the small noise limit, Levy-driven gradient descent dynamics need only polynomial time. This is in contrast to Brownian-motion driven gradient descent dynamics, which need exponential time to transit to another minimum \cite{simsekli2019tail}.

Another consequence is that the mean transition time between metastable valleys under Lévy-driven SDEs remains independent of the potential well depth $H$. This stands in contrast to Brownian motion-driven systems, where escape timescales grow significantly with increasing $H$. This can be explained by the fact that Brownian systems must overcome the potential barrier through gradual ascent, whereas Lévy systems can escape through discontinuous jumps, bypassing the need to climb gradients. This property makes Lévy-driven dynamics particularly robust in energy landscapes containing deep potential wells.

And all these features of Lévy-driven dynamics are tightly related to parameter $\alpha$ (whose limit $\alpha\rightarrow 2$ corresponds to Brownian motion dynamics). In this context, numerically quantifying such parameter becomes paramount, both to understand the internal behavior of SGD algorithm and to predict the impact of decisions like changes in the architecture and hyperparameters of a model trained by this method.

\subsection{Fourier-based optimizer}

To further leverage the potential of Fourier analysis on the process of deep neural networks generalization, here we also propose a customized optimizer. Our algorithm takes into account an indirect estimation of the Fourier measure developed here. More specifically, the Fourier dimension is computed over the learnable parameters tensor at each layer. 

\begin{figure}
    \centering
    \includegraphics[width=0.5\textwidth]{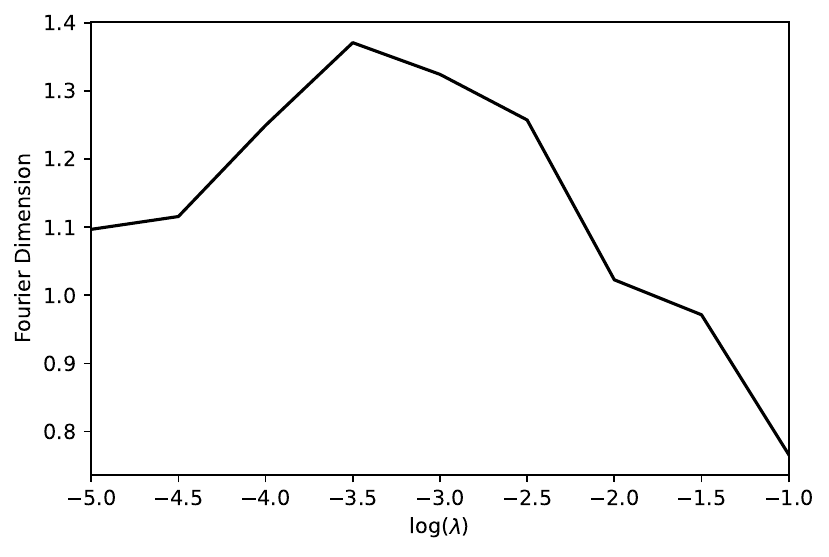}
    \caption{Impact of Fourier optimizer parameter on the proposed generalization measure.}
    \label{fig:optimizer_dimension}
\end{figure}

This is a feasible approximation to the computation of the dimension over the evolution of parameters across epochs. This last procedure would be impractical given that considering all the history of parameter values per epoch would require abrupt updates of parameter values to adjust the Fourier dimension at each epoch. The training process would become extremely unstable in this way. At the same time, our optimizer enforces general reduction on the magnitude of the Fourier transform of the parameters at each epoch, which at the end implies the reduction of the overall Fourier dimension, proposed here as our predictor for the generalization gap. Figure \ref{fig:optimizer_dimension} illustrates how the proposed optimizer also implies trained models with reduced Fourier generalization measure.

\section{Validation Setup}

To evaluate the effectiveness of the proposed Fourier fractal dimension as a generalization measure, we conducted extensive experiments on three standard image classification benchmarks: CIFAR-10, SVHN, and MNIST. The neural network architectures consisted of standard convolutional networks, including a modified AlexNet, adapted for the respective image resolutions and complexities. Models were trained utilizing both standard Stochastic Gradient Descent (SGD) and our proposed Fourier-based optimizer. During the training process, the learning rate was dynamically adapted using a plateau-based scheduler (e.g., \texttt{ReduceLROnPlateau}(mode='min', factor=0.1, patience=5)) to ensure steady convergence without outputting unnecessary verbose logging. All models were trained for a maximum of 100 epochs, actively recording the parameters' log-variations at each layer to compute the corresponding Lévy stable distribution fits and the Fourier dimensions.

\section{Results and Discussion}\label{sec:results}

The experimental results demonstrate a strong empirical correlation between our proposed Fourier generalization measure and the actual generalization gap. As detailed in Table \ref{tab:measures}, we compared the Kendall rank correlation coefficient ($\tau$) of our proposed metric against a wide array of existing measures documented in \cite{jiang2019fantastic}. Our method achieved the highest Kendall coefficients across all three datasets: $0.680$ on CIFAR-10, $0.672$ on SVHN, and $0.551$ on MNIST, significantly outperforming competitive baselines such as PAC-Bayes flatness and inverse margin.
\begin{table}[!htpb]
	\centering
	\caption{Kendall coefficients on CIFAR-10, SVHN, and MNIST databases for the proposed measure in comparison with the literature. The measures are defined in \cite{jiang2019fantastic}.}
	\label{tab:measures}
	\begin{tabular}{llll}
		\hline
		Measure & CIFAR-10 & SVHN & MNIST\\
		\hline
		$L_2$-norm & -0.419 & -0.200 & -0.307\\
		$L_2$-distance & -0.414 & -0.200 & -0.294\\
		\# of parameters & -0.048 & -0.468 & -0.327\\
		inverse margin & 0.473 & 0.657 & 0.386\\
		log-prod spectral & -0.315 & -0.238 & -0.333\\
		log-prod spectral margin & -0.256 & -0.200 & -0.333\\
		log spectral init & -0.266 & -0.209 & -0.333\\
		Frobenius spectral & 0.059 & 0.362 & -0.333\\
		log spectral origin & -0.266 & -0.209 & -0.333\\
		log-sum spectral margin & -0.256 & -0.200 & -0.333\\
		log-sum spectral & -0.315 & -0.238 & -0.333\\
		log-prod Frobenius & -0.438 & -0.152 & -0.268\\
		log-prod Frobenius margin & -0.429 & -0.124 & -0.242\\
		log-sum Frobenius margin & -0.429 & -0.124 & -0.242\\
		log-sum Frobenius & -0.438 & -0.152 & -0.268\\
		Frobenius distance & -0.414 & -0.200 & -0.294\\
		distance spectral init & -0.271 & 0.048 & -0.085\\
		parameter norm & -0.419 & -0.200 & -0.307\\
		path norm & -0.369 & 0.133 & -0.046\\
		path norm margin & -0.320 & 0.143 & -0.020\\
		PAC-Bayes init & 0.429 & 0.619 & 0.516\\
		PAC-Bayes origin & 0.369 & 0.676 & 0.516\\
		PAC-Bayes flatness & 0.613 & 0.545 & 0.546\\
		PAC-Bayes magnitude init & -0.172 & 0.171 & 0.111\\
		PAC-Bayes magnitude origin & -0.236 & 0.124 & 0.046\\
		PAC-Bayes magnitude flatness & 0.149 & 0.584 & 0.364\\
        Weight watcher & -0.201 & -0.326 & -0.218\\
		\hline
		Proposed & \textbf{0.680} & \textbf{0.672} & \textbf{0.551}\\
		\hline
	\end{tabular}
\end{table}

\begin{figure}[!htpb]
    \centering
    \begin{tabular}{c}
        \begin{tabular}{cc}
            CIFAR-10 & SVHN\\
            \includegraphics[width=0.5\linewidth]{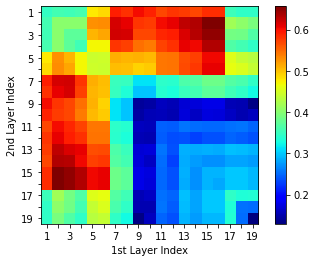} & \includegraphics[width=0.5\linewidth]{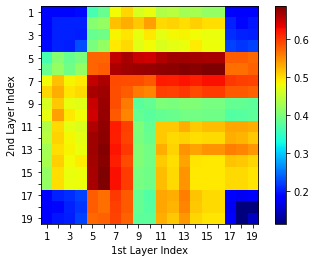} \\
        \end{tabular}   
        \\
        MNIST\\
        \includegraphics[width=0.5\linewidth]{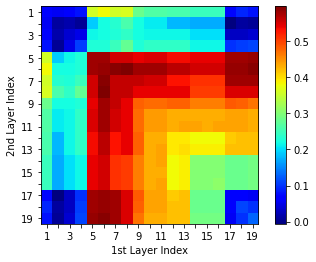}
    \end{tabular}
    \caption{Kendall's coefficients at different layer combinations.}
    \label{fig:tau_matrix}
\end{figure}

Figure \ref{fig:tau_matrix} illustrates the Kendall's coefficients calculated at different layer combinations for the three datasets. The heatmaps reveal that deeper layers tend to exhibit stronger correlations with the generalization gap, corroborating the hypothesis that higher-level semantic representations are critical for model generalizability. 

\begin{figure}[!htpb]
    \centering
    \begin{tabular}{cc}
        \includegraphics[width=0.5\textwidth]{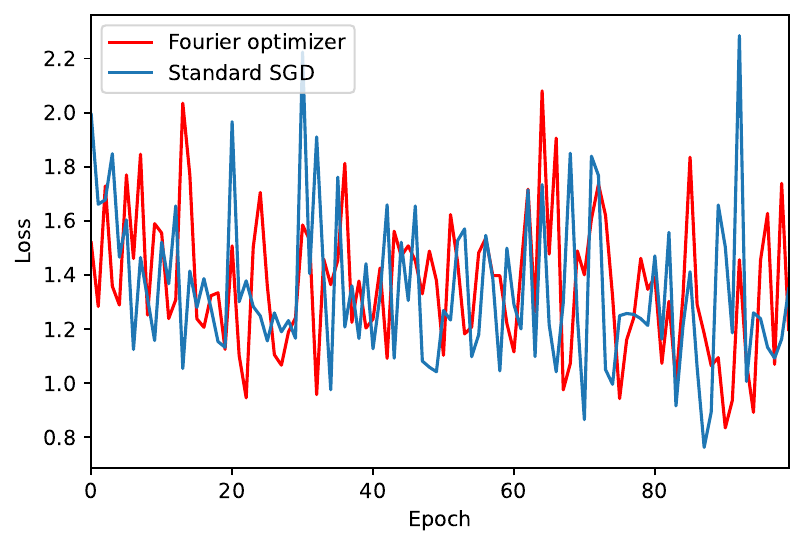} &  
        \includegraphics[width=0.5\textwidth]{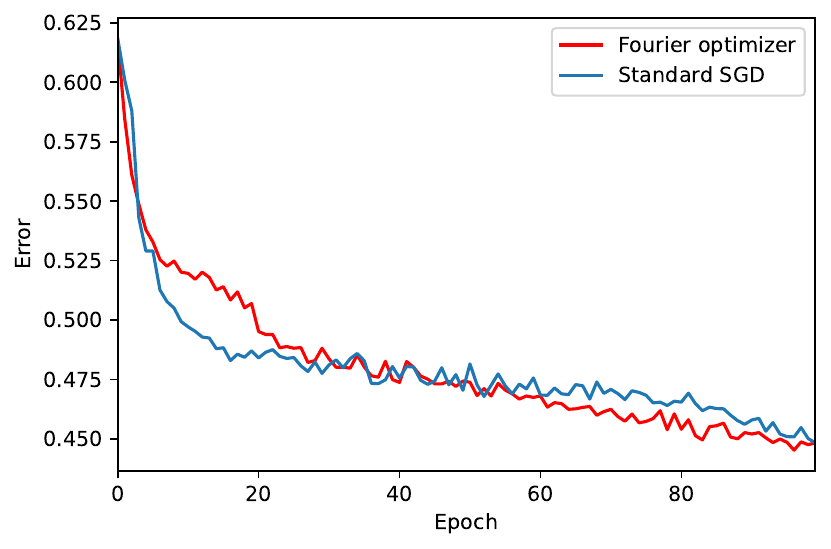}
    \end{tabular}
    \caption{Comparison of loss function and accuracy using vanilla stochastic gradient descent and the proposed Fourier optimizer.}
    \label{fig:optimizer}
\end{figure}

Furthermore, Figure \ref{fig:optimizer} compares the training dynamics of standard SGD against our proposed Fourier optimizer. The Fourier optimizer not only maintains a highly competitive error rate but also smooths the loss landscape over the epochs, yielding a more stable convergence profile. This stability clearly indicates that actively regulating the Fourier dimension during training can effectively mitigate extreme parameter oscillations.

\section{Conclusions}

In this work, we introduced a novel framework for predicting the generalization gap of deep neural networks based on the Fourier fractal dimension of the training dynamics. By modeling the Stochastic Gradient Descent as a Lévy-driven stochastic differential equation, we extracted the scale-invariant properties of the parameter updates. Our empirical validation across multiple benchmark datasets confirmed that the proposed Fourier dimension exhibits a remarkable correlation with the generalization gap, outperforming traditional norm-based and PAC-Bayesian measures. Additionally, the introduction of a Fourier-based optimizer highlights the practical utility of this metric, offering a principled approach to enhancing training stability and model generalization purely from internal training dynamics, without relying on external validation datasets.

\section*{Acknowledgements}

J. B. Florindo gratefully acknowledges FAEPEX/UNICAMP (Grant \#2493/23), S\~ao Paulo Research Foundation (FAPESP) (Grant \#2024/01245-1), and National Council for Scientific and Technological Development, Brazil (CNPq) (Grant \#306981/2022-0), for funding this research. 

\bibliographystyle{elsarticle-num} 
\bibliography{FourierGeneralization}

\end{document}